\documentclass[journal, twocolumn,final]{IEEEtran}
\usepackage{amsfonts}
\usepackage{amssymb}
\usepackage{amsmath}
\usepackage{dsfont}
\usepackage{graphicx}
\usepackage{bm}
\usepackage{cite}
\usepackage{bigstrut}
\usepackage{float}
\usepackage{balance}
\usepackage{lipsum}
\usepackage{psfrag}
\usepackage{xcolor}
\usepackage{xfrac}
\usepackage{hyperref}
\usepackage[linesnumbered,ruled,vlined]{algorithm2e}
\usepackage{tikz}
\usepackage{pgfplots}
\pgfplotsset{compat=1.15}
\usepackage{mathrsfs}
\usetikzlibrary{arrows}
\usetikzlibrary{decorations.pathreplacing}
\usetikzlibrary{fadings}

\newtheorem{definition}{Definition}

\newtheorem{lemma}{Lemma}

\usepackage{mathtools}
\SetKwInput{KwInput}{Input}                
\SetKwInput{KwOutput}{Output}

\begin{document}
\IEEEoverridecommandlockouts
\allowdisplaybreaks

\title{Split Learning in Computer Vision for Semantic Segmentation Delay Minimization}

\author{Nikos G. Evgenidis, Nikos A. Mitsiou,~\IEEEmembership{Graduate Student Member,~IEEE}, Sotiris A. Tegos,~\IEEEmembership{Senior Member,~IEEE}, Panagiotis D. Diamantoulakis,~\IEEEmembership{Senior Member,~IEEE}, and George K. Karagiannidis,~\IEEEmembership{Fellow,~IEEE}
\thanks{The authors are with the Department of Electrical and Computer Engineering, Aristotle University of Thessaloniki, 54124 Thessaloniki, Greece (e-mails: nevgenid@ece.auth.gr, nmitsiou@auth.gr, tegosoti@auth.gr padiaman@auth.gr, geokarag@auth.gr).}
}
\maketitle


\begin{abstract}
In this paper, we propose a novel approach to minimize  the inference delay in semantic segmentation using split learning (SL), tailored to the needs of real-time computer vision (CV) applications for resource-constrained devices. Semantic segmentation is essential for applications such as autonomous vehicles and smart city infrastructure, but faces significant latency challenges due to high computational and communication loads. Traditional centralized processing methods are inefficient for such scenarios, often resulting in unacceptable inference delays. SL offers a promising alternative by partitioning deep neural networks (DNNs) between edge devices and a central server, enabling localized data processing and reducing the amount of data required for transmission. Our contribution includes the joint optimization of bandwidth allocation, cut layer selection of the edge devices' DNN, and the central server's processing resource allocation. We investigate both parallel and serial data processing scenarios and propose low-complexity heuristic solutions that maintain near-optimal performance while reducing computational requirements. Numerical results show that our approach effectively reduces inference delay, demonstrating the potential of SL for improving real-time CV applications in dynamic, resource-constrained environments.
\end{abstract}
\begin{IEEEkeywords}
Computer vision, semantic segmentation, split learning, inference delay
\end{IEEEkeywords}
\section{Introduction} \label{Sec:Intro}

\IEEEPARstart{R}{eal}-time computer vision (CV) has become essential for a variety of emerging applications, including digital manufacturing, advanced transportation, remote sensing, robotics, virtual/augmented reality (VR/AR), and the metaverse \cite{nalbant2021computer}. Semantic segmentation, a specific CV technique, plays a crucial role in these areas by labeling each pixel in an image with a class label to partition the image into semantically meaningful segments, providing a highly detailed understanding of scenes \cite{rts}. Such granular segmentation is critical for accelerating decision making in dynamic environments, making it invaluable for real-time CV-based applications. 
To this end, efficient deep neural networks (DNNs) have accelerated real-time processing, making semantic segmentation a viable option for latency-sensitive applications. Convolutional neural network (CNN) architectures such as U-Net \cite{Ronneberger2015}, ENet \cite{Paszke2016}, and DeepLab \cite{Chen2018} have been developed to optimize both accuracy and speed, allowing them to process and analyze complex scenes with high accuracy. Moreover, recent innovations in efficient DNNs focus on reducing computational complexity and inference time, which are key to deployment in resource-constrained environments such as edge devices and internet-of-things (IoT) networks. 

Despite these advances, real-time CV also faces significant challenges due to limited communication resources, as large amounts of data must be transmitted across distributed entities. Therefore, joint optimization of communication and computational resources is essential to enable latency-efficient CV-based decisions \cite{tam2021adaptive}. Emerging machine learning (ML) paradigms that integrate distributed computing and edge intelligence, such as split learning (SL) \cite{hafi2024split}, are a promising option for such a holistic real-time CV solution \cite{siddique2021towards}. Instead of transmitting entire datasets or high-resolution images to a central server, SL divides a DNN model between edge devices and a central server, allowing each segment of the model to process its portion of the data locally before sharing only the necessary intermediate outputs \cite{proceedings}. This approach significantly reduces the amount of data transmitted, while allowing the slicing layer to be dynamically adjusted based on channel conditions, device power, and computing resources. 

\subsection{State of the Art}
For real-time CV applications on resource-constrained devices, efficient DNN models have been developed. In \cite{Shelhamer2017}, fully convolutional networks replaced fully connected layers, allowing end-to-end training for pixel-wise classification and supporting inputs of different sizes. This innovation set the stage for subsequent models, such as U-Net \cite{Ronneberger2015}, which used a symmetric encoder-decoder structure with skip connections to combine low and high-level features, improving detail representation and contextual understanding. Furthermore, SegNet \cite{Badrinarayanan2017} utilized encoder pooling indices to minimize additional parameters, while ENet \cite{Paszke2016} employed early downsampling and factorized convolutions to enable faster inference with minimal loss of accuracy. Improvements in segmentation accuracy were also made in DeepLab \cite{Chen2018}, which used atrous convolutions and conditional random fields to capture multi-scale context and improve boundary precision. DeepLab evolved into DeepLabv3+ for further refinement \cite{Chen2018a}. Other models, such as RefineNet \cite{Lin2017} and ERFNet \cite{Romera2018}, emphasize efficiency and precision, with RefineNet preserving high-resolution details and ERFNet integrating residual connections and factorized convolutions for real-time performance, making it suitable for autonomous driving. 

Moreover, to further optimize the performance of CV-based applications, their integration with distributed ML architectures has been investigated. In particular, the integration of federated learning (FL) has been proposed to accelerate distributed collaborative learning of CV tasks. For example, \cite{heetall} developed FedCV, a versatile FL framework addressing data heterogeneity and privacy across different tasks, while \cite{fl2} highlighted the suitability of FL for segmentation with minimal communication requirements. In addition, \cite{fl3} demonstrated FedVision, an FL-based object detection platform that operates without centralizing sensitive data, illustrating practical industrial applications of FL, while \cite{fl4} applied FL to vision-language tasks, improving model collaboration for tasks such as image captioning. However, FL cannot be used to accelerate the inference delay of an ML model once it has been trained.

SL is a distributed ML paradigm that can address this challenge. Its growing adoption in both academia and industry includes implementation in open-source applications \cite{sl1, sl2} and services from startups \cite{sl3}. Academic research encompasses empirical studies in various fields, e.g., \cite{sl4} applied SL to depth-image-based millimeter-wave power prediction and achieved significant latency reductions, while other studies explored SL in medical imaging \cite{sl5}. Comparative studies also highlighted the reduced communication overhead of SL compared to FL \cite{sl6}. Research on SL privacy showed that it provides higher protection than FL since devices do not have access to server-side models, as noted in \cite{sl7}. Further research optimizes SL by introducing variants with multiple cut layers, allowing devices to retain initial and final layers to protect data and labels \cite{sl8}. Furthermore, joint optimization of communication and computation resources for SL was studied in \cite{sl9,sl10,sl11}, showing significant latency gains. However, the integration of SL with CV-based tasks has not yet been investigated in the literature. 

\subsection{Motivation and Contribution}
The interest in exploring the SL paradigm in semantic segmentation and CV lies in its potential to address the high computational and communication burden associated with pixel-level predictions, especially in resource-constrained devices. Semantic segmentation, while critical for applications such as autonomous vehicles and smart city infrastructure, requires intensive computation due to the need to make dense, per-pixel predictions. As a result, traditional centralized approaches, where an entire model runs on either an edge device or a central server, can result in significant communication and computation latency, often making real-time processing infeasible. Therefore, to fully realize the potential of SL in CV-based applications, a joint optimization of computational and communication resources is required. This includes strategies for partitioning the CV models between the edge devices and the central server of the network according to their computational capabilities and the data load of the central server, and optimizing the data transmission between the devices and the server through optimal bandwidth allocation.

To address these challenges, this paper aims to minimize the inference delay in semantic segmentation by adopting a SL approach. Unlike SL for feedforward neural networks (FNNs) in \cite{sl9, sl10, sl11}, SL for CV tasks must accommodate the complexity of multi-layer CNN architectures and their reliance on \emph{bottleneck modules (BMs)}. These modules, which are integral to CNNs, do not allow splitting within a BM, unlike the simpler successive layers in FNNs. In addition, the data size to be transmitted when splitting such a DNN at a given layer is significantly different from that in FNNs, as will be explained in the next section. 
Furthermore, we consider two different use cases. First, the case where the central server processes all incoming tasks in parallel, and second, the case where tasks are processed sequentially. Each scenario leads to different optimal layer-slicing policies and resource allocation strategies.
Moreover, to facilitate real-time implementation, we propose heuristic, sub-optimal layer-cutting and resource allocation policies that offer significantly lower complexity compared to the optimal delay minimization scheme. The contribution of the paper is given below:

\begin{itemize}
    \item For the first time in the literature, we propose the use of SL to minimize the inference delay in CV. To achieve this, we take into account the special features of CV, such as multi-layer CNNs and BM modules, which complicate optimal layer splitting.
    \item For two cases of interest, the case of data processed in parallel by the central server and the case of serial data processing, the problem of joint bandwidth allocation, cut layer selection of the edge devices' DNN, and central server processing resource allocation is tackled. The initial problem is non-convex, but by using alternating optimization, it is reduced to a convex optimization sub-problem and a binary programming sub-problem, which can be solved efficiently.
    \item To support real-time CV, low-complexity algorithms are provided by optimizing only a subset of the variables. Specifically, for the parallel data processing case, by fixing the bandwidth allocation, an equivalent formulation of the problem with a closed-form solution is given, while a scheme that uses the cutting layer with the least amount of data to be transmitted is also studied. For the case of serial data processing, a strategy based on the simultaneous arrival of data at the central server is investigated. Moreover, a low-complexity variant is studied, which reallocates the bandwidth among the devices appropriately, aiming to minimize their maximum waiting time in the queue upon arrival.
    \item Numerical results provide useful insights into the effectiveness of SL and all considered schemes. It is shown that SL can reduce the inference delay of CV applications, while trade-offs between all proposed schemes are revealed, which can be used to select the most appropriate algorithm given the available communication and computational resources and the number of devices. 
\end{itemize}

\subsection{Structure}
The rest of the paper is organized as follows. Section \ref{sec:SysMod} presents the system model. Section \ref{sec:pol} introduces the proposed slicing policy. In Section \ref{sec:DelayInf}, we formulate and solve the inference delay minimization problem. Finally, Section \ref{sec:num} presents the numerical results and Section \ref{sec:concl} concludes the~paper.

\begin{figure}
\centering
\includegraphics[width=0.98\columnwidth]{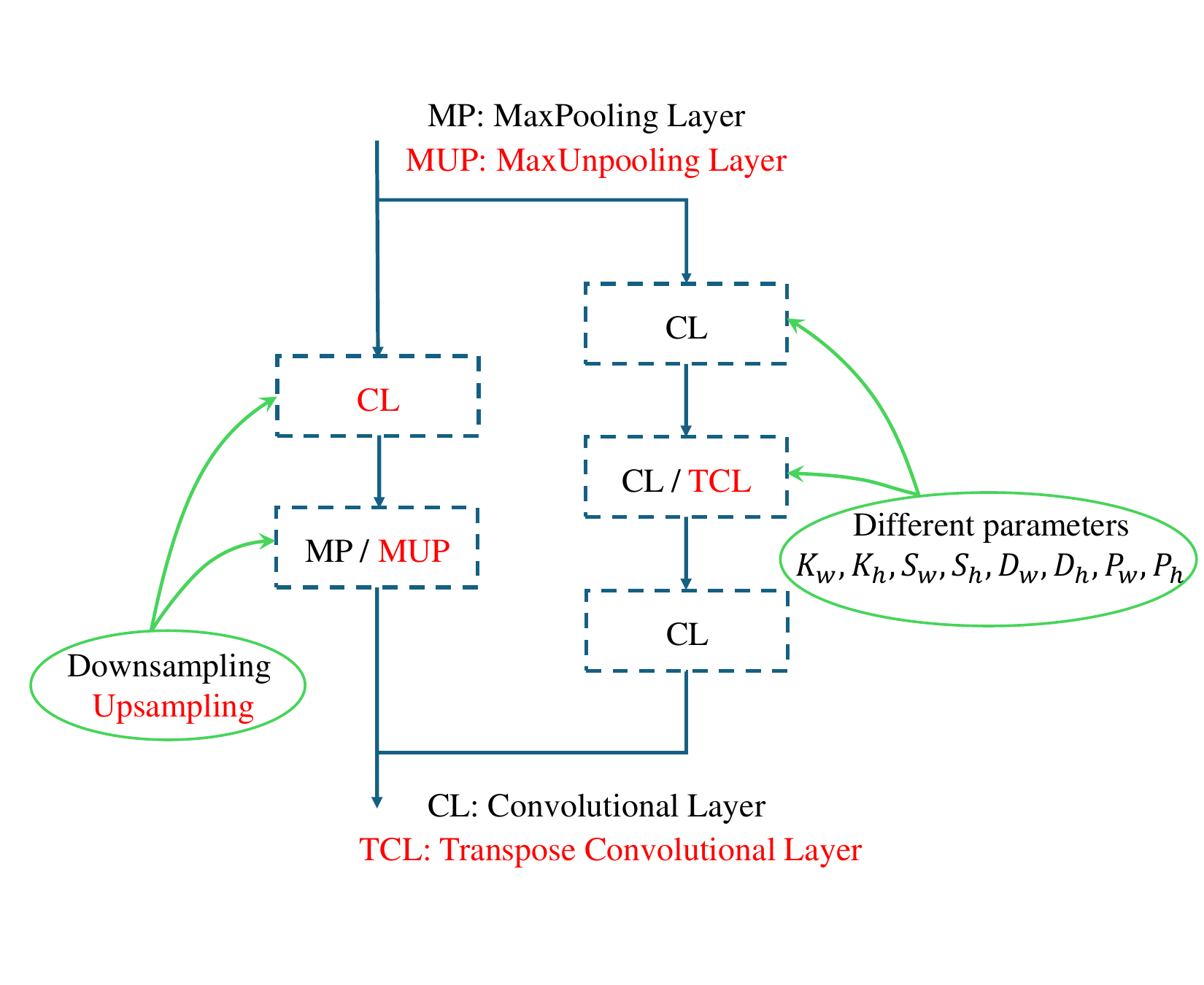}
\vspace{-1cm}
\caption{Typical design of a BM.}
\label{fig:BM}
\end{figure}

\section{System Model} \label{sec:SysMod}
Many current CV applications, including semantic segmentation, rely heavily on complex multi-layer CNN architectures. Although the cornerstone of these frameworks are convolutional layers, in order to enhance feature extraction, the specific frameworks implement the so-called BM instead of simple consecutive layers. The main characteristic of the BM is that it acts on two separate branches, which are added just before the output of the BM, as shown in Fig. \ref{fig:BM}. In addition, each BM inner structure can be different from others, depending on its type. This means that some of its layers can have different parameters, which are given in Table \ref{tab:Params}. This variety of inner structures ensures the capture of different characteristics, since each parameter affects the effective area of the included elements as the kernel slides along the data. Therefore, each BM can consist of multiple layers on each branch, which increases the workload of the CNNs and does not allow splitting while inside a BM. Obviously, the floating point operations per second (FLOPS) workload of each BM consists of that of multiple CNNs and pooling workloads and can be calculated using the formulas in Table \ref{tab:Works}. Although BMs are quite common in CNN architectures, semantic segmentation CV frameworks have unique characteristics that differentiate them from others.

\renewcommand{\arraystretch}{1.4}
\begin{table}
    \centering
    \caption{Parameters of convolutional and pooling layers.}
    \begin{tabular}{c||c}
    \hline \hline
    Channel inputs / outputs & $C_{in}$ / $C_{out}$ \\
    \hline \hline
    Input width / height & $W_{in}$ / $H_{in}$ \\
    \hline
    Output width / height & $W_{out}$ / $H_{out}$ \\
    \hline
    Kernel width / height & $K_w$ / $K_h$ \\
    \hline
    Padding width / height & $P_w$ / $P_h$ \\
    \hline
    Stride width / height & $S_w$ / $S_h$ \\
    \hline
    Dilation width / height & $D_w$ / $D_h$ \\
    \hline
    Output padding width / height & $P_{wo}$ / $P_{ho}$ \\
    \hline \hline
    \end{tabular}
    \label{tab:Params}
\end{table}

\begin{table}
    \centering
    \caption{Output size $X_{out}$ ($X=H,W$) and workload $O$ (FLOPS) for each type of layer.}
    \begin{tabular}{c}    
    \hline \hline
    {\textbf{Convolutional Layer}} \\
    \hline
    $X_{out} = \left\lfloor \frac{X_{in} + 2P_x - D_x(K_x - 1) - 1}{S_x} + 1 \right\rfloor$ \\
    \hline
    $O = 2 C_{in} C_{out} K_w K_h W_{out} H_{out}$ \\
    \hline
    \hline
    {\textbf{Transpose Convolutional Layer}} \\
    \hline
    $X_{out} = (X_{in} - 1)S_x - 2P_x + D_x(K_x - 1) + P_{xo} + 1$ \\
    \hline
    $O = 2 C_{in} C_{out} K_w K_h W_{out} H_{out}$ \\
    \hline    
    \hline
    {\textbf{MaxPooling Layer}} \\
    \hline
    $X_{out} = \left\lfloor \frac{X_{in} + 2P_x - D_x(K_x - 1) - 1}{S_x} + 1 \right\rfloor$ \\
    \hline
    $O = (K_w K_h -1) C_{out} W_{out} H_{out}$ \\
    \hline
    \hline
    {\textbf{MaxUnpooling Layer}} \\
    \hline
    $X_{out} = (X_{in} - 1)S_x - 2P_x + K_x$ \\
    \hline
    \rule{1cm}{0.1mm} \\
    \hline \hline
    \end{tabular}
    \label{tab:Works}
\end{table}

\begin{figure*}
\centering
\includegraphics[width=1\linewidth]{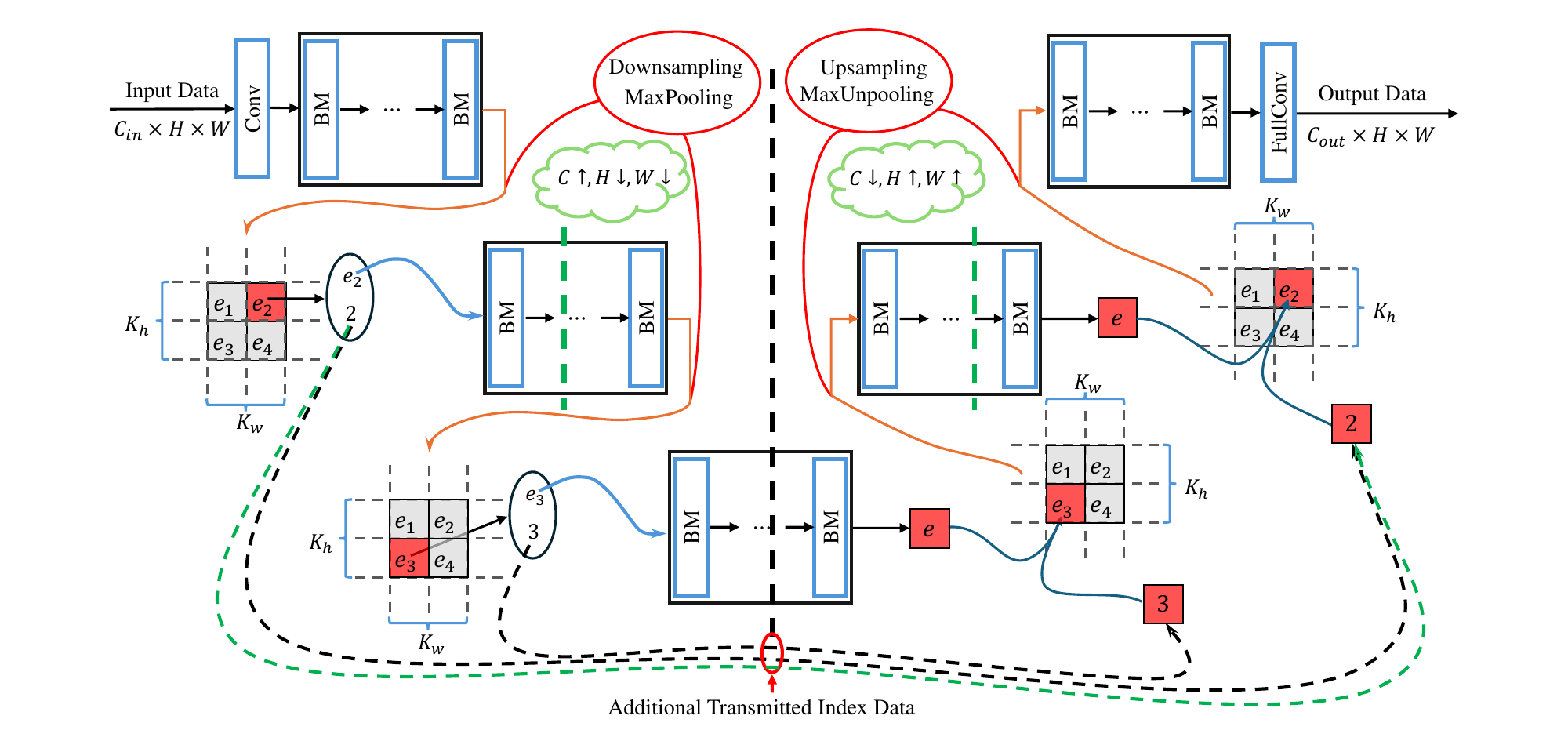}
\caption{A CV semantic segmentation DNN architecture consisting of multiple stacks of BMs of different characteristics.}
\label{fig:CV_arch}
\end{figure*}


A distinctive feature of semantic segmentation frameworks is downsampling and upsampling of image dimensions, which is achieved by changing the number of filters for better feature extraction. The downsampling process can be performed in two ways, either by using pooling layers or convolutional layers with stride, and requires a specific inner layer structure in its BM as shown in Fig. \ref{fig:BM}. On the other hand, to obtain the classification of each pixel of the original image, these architectures also require upsampling, which is performed in a similar way to the downsampling part, as shown in Fig. \ref{fig:CV_arch}. This means that if a pooling layer was used, an unpooling layer must be used for upsampling, and if a convolutional strided layer was used, a corresponding convolutional transpose layer should be used for upsampling. Note that downsmapling/upsampling is generally performed over MaxPooling/MaxUnpooling layers, except for the first and last layers, which use convolutional layers depicted as ``Conv" and ``FullConv" in Fig. \ref{fig:CV_arch}. Also, since upsampling is the inverse of downsampling, the number of upsampling layers must be equal to the number of downsampling layers to ensure that the output of the architecture has the same dimensional size with respect to the original height and width of the image. Regarding the data output of each BM in a stack, it is important to emphasize that it remains the same compared to the input of the BM, since the number of input and output filters, as well as the height and width, are preserved, unless a BM is used for downsampling or upsampling, in which case it is decreased or increased, respectively. 

In addition, unpooling layers require not only the data to perform upsampling, but also the corresponding pooling layer index. Thus, as shown in Fig. \ref{fig:CV_arch}, an upsampling BM requires the processed data and the pooling layer index, which indicates where in the unpooling kernel the output-element, denoted as $e$ in Fig. \ref{fig:CV_arch}, of the upsampling BM will be placed. For example, in Fig. \ref{fig:CV_arch} a MaxPooling is performed with $K_h\!=\!K_w\!=\!2$, meaning that the maximum of each $K_hK_w$ elements is forwarded to the BMs for further processing. Moreover, the index of this element is used for its MaxUnpooling counterpart to create a frame of $K_hK_w$ elements, in which the BM's output-element is placed, while the rest of the elements are filled using an interpolation technique between neighboring elements. Therefore, some BMs may require an additional amount of data to be generated for later use when downsampling is performed. Note that this data can be relatively small compared to the processed data, e.g., a downsampling BM with MaxPooling layer with $K_h\!=\!K_w\!=\!2$ requires only $2$ bits to keep the index information of each kernel, as shown in Fig. \ref{fig:CV_arch}. Apart from this, the index data is necessary for correct upsampling and thus provides some protection against potential leakage of the processed data of the framework.

In this paper, we are interested in the timely inference of all devices in a communication network. For this purpose, we consider a network consisting of $K$ devices communicating with a single base station (BS) where the central server is also located. For the communication of each device, we assume that the transmit power of the $k$-th device is constant and equal to $P_k$, while the available bandwidth of the $k$-th device is denoted as $B_k$. Considering the small-scale fading coefficient $h_{k}$ of the channel between each device and the BS, the maximum achievable rate is given by 
\begin{equation}
    R_k = B_k \log_{2} \left(1 + \frac{P_k G_t G_r L_p(d) |h_k|^2}{N_0B_k} \right),
\end{equation}
where $N_0$ is the noise power spectral density, $G_t$ and $G_r$ are the transmitter and receiver gains, respectively, while 
$L_p(d) = \left(4 \pi d / \lambda \right)^{-n}$ is the path loss at a transmit frequency corresponding to wavelength $\lambda$ at distance $d$ from the BS, and $n$ is the path loss exponent, which varies with the environment. With respect to the computational capabilities of the transmitters, let $f_k$ be the computational resources of the $k$-th device.

\section{Slicing Policy for Semantic Segmentation} \label{sec:pol}
One of the main requirements of many real-time applications in modern communication networks is a low inference delay to ensure normal functionality. Consequently, in CV applications such as semantic segmentation, it is critical for the system to minimize the total inference time of all participating devices. With this in mind, we propose an SL-inspired \textit{slicing policy} that allows all devices to ``slice'' their used architecture at any BM and transmit the necessary data for further processing at the central server, thus fully exploiting the processing capabilities of both communication ends and aiming to decrease inference time.

The rationale behind the proposed policy is that the two extreme cases, i.e., transmitting raw pixel data and transmitting the pixel-wise prediction output of the last layer of the semantic segmentation architecture, can both lead to increased delay. In the first case, the central server of the BS must simultaneously process the entirety of the generated data in parallel, which increases the inference delay, while in the second case, the inference delay is increased due to the  limited computational resources at each device. Furthermore, unlike other CNN frameworks, where the amount of data per layer is usually consistently reduced, a typical semantic segmentation architecture, such as the one shown in Fig. \ref{fig:CV_arch}, may have more data to process than previous BMs when upsampling takes place. Therefore, there is a trade-off between processing and communication delay that can be addressed by the proposed SL-based slicing policy.

In this context, we assume that each device implements its own semantic segmentation architecture consisting of $L$ BMs. Then, the processing time of the $l$-th BM module at the $k$-th device is denoted as $W_{k,l}$ and the total processing time of a framework results from the summation of all individual workloads. Except for the workload $W_{k,l}$ of each BM module, each of the latter generates data that must be passed to the next module, which is denoted as $D_{k,l}$ for the $k$-th device at the $l$-th BM. Then, if the $k$-th device decides that it is advantageous to execute the \textit{slicing policy} at the $l$-th BM, $D_{k,l}$ must be transmitted at rate $R_{k}$. Unique to this type of architecture, an additional amount of required data, denoted as $\tau_{k,l}$ for the $k$-th device slicing at the $l$-th BM, may also need to be transmitted if upsampling must be performed at the BS, as explained in Section \ref{sec:SysMod}. It should be noted that this amount will be different depending on the sliced BM. For example, as shown in Fig. \ref{fig:CV_arch}, if a device chooses to slice at the black dashed line, then it must transmit index-related data from two downsampling BMs, while if it slices at one of the two green dashed lines, index-related information from only one BM is required. 

It is important to emphasize that, as explained in Section \ref{sec:SysMod}, the amount of $\tau_{k,l}$ is small compared to $D_{k,l}$, but still affects the overall communication delay and is required to complete processing at the BS. It is also important to emphasize that even if leakage of $D_{k,l}$ occurs, without knowledge of the corresponding $\tau_{k,l}$, a potential attacker will not be able to make correct inferences from the processed data. Therefore, the proposed \textit{slicing policy} also provides some real-time protection against data leakage due to the nature of the utilized MaxUnpooling layers in upsampling \cite{Paszke2016}, when the devices choose to transmit data before the final output.
Moreover, for the central server to successfully implement the proposed policy, we assume that it contains all possible instances of the devices' DNN architectures and can process data from any BM. Let $f_{k,s}$ denote the computational resources of the central server associated with the $k$-th device. For completeness, we consider two resource allocation schemes at the BS side, one where the resources are distributed among all devices and cannot reach values greater than a maximum computational limit, $f_{\mathrm{max}}$, and another where full resources are given to each device.

\section{Inference Delay Minimization}\label{sec:DelayInf}
The goal of this section is to minimize the inference delay of all devices, which contains the local processing delay, the transmission delay and the processing delay at the server. To this end, each device chooses to slice its DNN at a specific BM. To model the slicing layer selection, binary variables $a_{k,l}$ are introduced for the $k$-th device and its $l$-th BM layer. Obviously, only one slicing layer can be selected for each device, thus $\sum_{l=1}^{L} a_{k,l} = 1,\ \forall k \in \mathcal{K}$. Also, each device can control its bandwidth by dynamically allocating its spectrum from a total available bandwidth $B_{\mathrm{tot}}$, leading to $\sum_{k=1}^{K} B_k = B_{\mathrm{tot}}$. Next, we investigate the delay minimization problem for two possible scenarios, related to the way that the server processes the received data. 

\subsection{Parallel Processing}
From the above constraints, if the server chooses to process data from different devices in parallel, the inference delay of the $k$-th device sliced at the $l$-th BM is given as 
\begin{equation}\label{eq:inferenceTime}
    J_{k,l} = \frac{D_{k,l}}{R_k} + \frac{\tau_{k,l}}{R_k} + \frac{\sum_{j=1}^{l} W_{k,j}}{f_k} + \frac{\sum_{j=l+1}^{L} W_{k,j}}{f_{k,s}},
\end{equation} 
where the last part of \eqref{eq:inferenceTime} indicates a dynamic resource allocation at the central server, which means that parallel processing of data is performed. Taking \eqref{eq:inferenceTime} into account, we can now formulate the following general optimization problem 
 \begin{equation*}\tag{\textbf{P1}}\label{eq:OptProblemGeneral}
    \begin{array}{cl}
    \mathop{\mathrm{min}}\limits_{a_{k,l},f_{k,s},B_k}
    & \max \left\{ \sum\limits_{l=1}^{L} a_{k,l}J_{k,l} \right\} \\
    \text{\textbf{s.t.}}& \mathrm{C}_1: \sum\limits_{k=1}^{K} f_{k,s} = f_{\mathrm{max}}, \\
    & \mathrm{C}_2:  \sum\limits_{k=1}^{K} B_k = B_{\mathrm{tot}}, \\
    & \mathrm{C}_3:  \sum\limits_{l=1}^{L} a_{k,l} = 1, \,\,\, \forall l \in \{1, \cdots ,L \}.
    \end{array}
\end{equation*}

Obviously, \eqref{eq:OptProblemGeneral} cannot be solved directly because it involves binary and continuous variables. However, it can be observed that $J_{k,l}$ in \eqref{eq:inferenceTime} is convex in both $f_{k,s}$ and $B_k$. Therefore, for fixed $a_{k,l}$, the objective of \eqref{eq:OptProblemGeneral} is also convex as the maximum of convex functions. As a result, alternating optimization can be used to solve \eqref{eq:OptProblemGeneral} by iteratively solving a convex and a binary programming problem. For each device, an initial random slicing layer selection is made and a convex optimization problem is solved to extract the optimal $f_{k,s},B_{k}$. The obtained solution is then used to solve a binary programming problem in terms of $a_{k,l}$. 
Although integer programming techniques can effectively solve this subproblem, the required complexity can sometimes be large due to the large number of variables. Observing that for fixed $f_{k,s},B_{k}$ also fixes $J_{k,l}$, we can decompose the binary programming problem into $K$ independent minimization problems, each of which is equivalent to finding the minimum value of $J_{k,l},\ \forall k \in \mathcal{K}$, whose overall complexity is very small. Repeating this process until the iterative procedure converges yields a solution to the original general problem.

\subsubsection*{Special Case of Fixed Bandwidth}
Although \eqref{eq:OptProblemGeneral} can be efficiently solved, its complexity may be high for real time applications, especially as the number of devices in the system increases. Consequently, a simpler version of this problem that yields a near-optimal solution, but with considerably lower complexity, is of great interest. As such, by fixing the bandwidth allocation for all devices, the following optimization problem is formulated
\begin{equation*}\tag{\textbf{P2}}\label{eq:OptProblemSimple}
    \begin{array}{cl}
    \mathop{\mathrm{min}}\limits_{a_{k,l},f_{k,s}}
    & \max \left\{ \sum\limits_{l=1}^{L} a_{k,l}J_{k,l} \right\} \\
    \text{\textbf{s.t.}}& \mathrm{C}_1: \sum\limits_{k=1}^{K} f_{k,s} = f_{\mathrm{max}}, \\
    & \mathrm{C}_2:  \sum\limits_{l=1}^{L} a_{k,l} = 1, \,\,\, \forall l \in \{1, \cdots ,L \},
    \end{array}
\end{equation*}
which can be solved by the same method. However, instead of solving a convex optimization problem, we can equivalently transform the problem into a simpler form with much lower complexity. For any realization of $a_{k,l}$, the local processing and transmission delay of the device can be written as 
\begin{equation}\label{eq:Constants}
    C_{k} = \frac{D_{k,l} + \tau_{k,l}}{R_k} + \frac{\sum_{j=1}^{l} W_{k,j}}{f_k},
\end{equation}
while the inference workload at the server corresponding to this device is given as
\begin{equation}\label{eq:ConstantsF}
     F_{k} = \sum_{j=l+1}^{L} W_{k,j}.
\end{equation}
Then, to simplify \eqref{eq:OptProblemSimple}, we observe that a $\mathrm{min} \mathrm{max}$ problem reaches its optimal solution when there is inference time equality among all participating devices. Based on \eqref{eq:Constants} and \eqref{eq:ConstantsF}, for any two devices $k,m$ we must have
\begin{equation}\label{eq:equalCons}
    C_k + \frac{F_k}{f_{k,s}} = C_m + \frac{F_m}{f_{m,s}} \Leftrightarrow f_{k,s} = \frac{F_k f_{m,s}}{F_m + f_{m,s} \Delta c_{m,k}},
\end{equation}
where $\Delta c_{m,k} = C_m - C_k$. Let $\mathcal{F} = \{ k | F_k \neq 0 \}$ denote the set of devices that have not performed all the processing at their end before transmitting data, in which case the central server must allocate resources for them. If a device has finished processing locally, i.e. $k \notin \mathcal{F}$, the server does not allocate resources, thus $f_{k,s} = 0$ and the right hand-side of \eqref{eq:equalCons} still holds. Then, summing for all devices in the system, we get
\begin{equation}\label{eq:equationClosed}
    f_{\mathrm{max}} = f_{m,s} + \sum_{\substack{k \in \mathcal{F} \\ 
    k \neq m}} \frac{f_{m,s}F_k}{F_m + f_{m,s} \Delta c_{m,k}}.
\end{equation}

Clearly, \eqref{eq:equationClosed} has multiple roots in terms of $f_{m,s}$, some of which may result in infeasible frequency allocation solutions, as the denominator in \eqref{eq:equalCons} can be negative due to $\Delta c_{m,k}$. Therefore, $m$ must be chosen carefully to ensure the correctness of the proposed technique. To ensure the feasibility of the obtained solution, we prove the following Lemma.

\begin{lemma} \label{lem:solFeas}
A solution of \eqref{eq:equationClosed} exists in the interval $ (0, f_b) $, where 
\begin{equation}\label{eq:conditions}
    f_b = \underset{{\substack{k \in \mathcal{F} \\ 
    k \neq m}}}{\mathrm{min}} \left\{ -\frac{F_m}{\Delta c_{m,k}} \right\} \hspace{0.5mm} \text{ and } \hspace{0.5mm} m = \underset{k \in \mathcal{F}}{\mathrm{argmin}} \{ C_k \},
\end{equation}
so that the optimal feasible resource allocation solution for all devices can always be obtained by \eqref{eq:equalCons}. 
\end{lemma}

\begin{IEEEproof}
    Let $m$ be described by \eqref{eq:conditions}, thus $\Delta c_{m,k} \leq 0,\ \forall k \in \mathcal{F}$. For the family of functions $g_{k}(x) = \frac{xF_k}{F_m + x\Delta c_{m,k}}$, it is easy to observe that all members are strictly increasing functions in their domains and each has a vertical asymptote at $x_k = - \frac{F_m}{\Delta c_{m,k}}$. Then, in the interval $(0, f_b)$, where $f_b$ is given by \eqref{eq:conditions}, the function 
    \begin{equation}\label{eq:func}
        q(x) = x + \sum_{\substack{k \in \mathcal{F} \\ 
    k \neq m}} g_{k}(x)
    \end{equation}
    is strictly increasing, and since $q(0)=0$ and $\lim_{x \rightarrow f_b^{-}} q(x) = \infty$, there is a unique point $x_0$ such that $q(x_0) = f_{\mathrm{max}}$.
    Note that if $m$ was different, then $\Delta c_{m,k} \leq 0,\ \forall k \in \mathcal{F}$ does not hold, and thus a negative root of \eqref{eq:equationClosed} may result, leading to an infeasible solution. 
    For the derived solution it is easy to check that 
    \begin{align}\label{eq:validity}
        F_m + x_0 \Delta c_{m,k} 
        &\geq F_m + f_b \Delta c_{m,k} \nonumber \\ 
        &= F_m \left(1 - \frac{|\Delta c_{m,k}|}{\underset{\substack{k \in \mathcal{F} \\ 
    k \neq m}}{\mathrm{max}} \{|\Delta c_{m,k} |\}} \right)
    \geq 0,
    \end{align}
    where we take advantage of $\Delta c_{m,k} \leq 0,\ \forall k \in \mathcal{F}$. Therefore, by \eqref{eq:equalCons} all frequency allocation resources are positive, proving their feasibility.
\end{IEEEproof} 
Using Lemma \eqref{lem:solFeas}, we can greatly reduce the complexity, since we are practically searching for a root in a small interval.

\subsection{Serial Processing}
In contrast to parallel processing, in this case the central server uses the maximum computational resources for each device and processes the incoming data serially one by one. For this reason, the central server utilizes a queue of $K$ positions. Assuming that a particular BM has been selected for the \textit{slicing policy}, the delay of each device upon arrival at the queue is given by $C_k$ in \eqref{eq:Constants} plus a $F_k/f_{\mathrm{max}}$ term corresponding to the task processing delay from the server resulting from \eqref{eq:ConstantsF}. However, a device may need to wait for previous tasks in the queue to finish before entering its processing phase. Note that tasks enter the queue in order of arrival, so devices are ordered in the queue according to their delay \( C_k \), with shorter delays placing data earlier in the queue.

Let \(\mathcal{K}'\) represent a permutation of the devices in \(\mathcal{K}\) that reflects the arrival order of their data. We define the total inference delay \(I_k\) for a device \(k \in \mathcal{K}'\) as follows
\begin{equation}\label{eq:orderInf}
    I_{k} = 
    \begin{cases}
         C_{1} + \frac{F_1}{f_{max}}, &k=1 \\
         \mathrm{max} \left\{I_{k-1}, C_{k} \right\} + \frac{F_k}{f_{max}} , &1 < k \leq K,
    \end{cases}  
\end{equation}
Minimizing the overall inference delay for all devices reduces to minimizing \(I_K\), the delay of the last device in the queue. However, since a particular arrival order in the queue, \(\mathcal{K}'\), depends on the delay of the devices, this makes it extremely difficult to optimize bandwidth allocation, since a different bandwidth allocation results in a different queue, and thus a different delay per device according to \eqref{eq:orderInf}. Furthermore, with \(K\) devices in the queue, there are \(K!\) possible queue configurations, making the problem highly complex, especially for real-time applications. This motivates the need for simpler, low-complexity approaches that can still perform effectively. Accordingly, we investigate two alternative schemes that are more tractable.

\subsubsection{Local Processing and Transmission Delay Minimization}
To avoid a potentially long transmission delay that would result in an increased total inference time, a useful scheme is to use bandwidth allocation to achieve simultaneous arrival of all tasks at the BS, resulting in a unique queue order of the devices. In this case, the inference delay minimization problem can be formulated as follows
\begin{equation}\tag{\textbf{P3}}\label{eq:minmaxDevice}
    \begin{array}{cl}
    \mathop{\mathrm{min}}\limits_{a_{k,l},B_k}
    & \max \left\{ \sum\limits_{l=1}^{L} a_{k,l}C_{k,l} \right\} + \sum\limits_{k=1}^{K} \frac{F_{k,l}}{f_{\mathrm{max}}}\\
    \text{\textbf{s.t.}}& \mathrm{C}_1:  \sum\limits_{k=1}^{K} B_k = B_{\mathrm{tot}}, \\
    & \mathrm{C}_2:  \sum\limits_{l=1}^{L} a_{k,l} = 1, \,\,\, \forall l \in \{1, \cdots ,L \},
    \end{array}
\end{equation}
where $C_{k,l}, F_{k,l}$ are the extended definitions of $C_k, F_k$ in \eqref{eq:Constants} and \eqref{eq:ConstantsF} to include the BM index $l$, which now changes. Note that the choice of $a_{k,l}$ affects $F_{k,l}$. Using alternating optimization between $a_{k,l}, B_k$, we can derive an optimal solution to \eqref{eq:minmaxDevice} by iteratively solving a convex optimization problem and searching for the $l^{*}$-th layer that minimizes all sequences $C_{k,l^{*}},\ \forall k \in \mathcal{K}$. 

\subsubsection{Equal Bandwidth Allocation}
In this scheme, we aim to leverage the different arrival rates of device data. If a device's data arrives while the previous data is still being processed, no additional delay is introduced, as described in \eqref{eq:orderInf}. However, if a device's data arrives after all previous data has been fully processed, the central server sits idle, increasing the total inference delay. To address this problem, we propose a heuristic algorithm designed to reduce delays when such idle gaps are likely to occur. We begin by introducing some key definitions about the queuing system and the inference delay, taking into account the arrival order of the data.

\begin{definition}\label{def:queues}
A ``break'' between the data arrivals of devices occurs when a device \( k \) experiences a delay, i.e., its data arrives after the processing of the previous device has been completed. This can be expressed as \( C_{k} = \max \{I_{k-1}, C_{k}\} \), for some \( k \in \mathcal{K}' \). If there are no breaks, i.e., each device's data arrives as soon as the previous one finishes processing, the queue is called ``unbroken'', and this is represented by the condition \( I_{k-1} = \max \{I_{k-1}, C_{k}\},\ \forall k \in \mathcal{K}' \). A queue is called ``\((M, \mathcal{M})\)-broken'' if it contains \( M \) consecutive unbroken sub-queues, where the set \(\mathcal{M} = \{ k \mid I_{k-1} < C_{k}, \, 1 < k \leq K \}\) specifies the indices where \(M - 1\) breaks occur in ascending order within the queue.
\end{definition}

According to definition \ref{def:queues}, for each device in an unbroken queue consisting of $Q$ devices, it holds that 
\begin{equation}\label{eq:unbrokenPerMember}
    \begin{aligned}
            & I_q = I_{q-1} + \frac{F_k}{f_{max}} ,\ \forall q \in \mathcal{Q},
            & I_{q_1} = C_{q_1} + F_{q_1}/f_{\mathrm{max}},
    \end{aligned}
\end{equation}
where $\mathcal{Q}$ is the set containing all $Q$ devices in the investigated unbroken queue, and $q_1$ is the index of the first device in this queue. Thus, the total inference delay for all devices in this queue is given by
\begin{equation}\label{eq:unbrokenQ}
    T_Q = I_Q = C_{q_1} + \sum\limits_{k\in\mathcal{Q}} \frac{F_k}{f_{max}}.
\end{equation}
Then, by definition, the total inference delay of the $(M,\mathcal{M})$-broken queue, denoted as $T_K^{(M,\mathcal{M})}$, is given by
\begin{equation}\label{eq:totBroken}
    T_K^{(M,\mathcal{M})} = I_K = C_{\mathrm{max} \{ \mathcal{M}\} } + \sum\limits_{k = \mathrm{max} \{ \mathcal{M}\}}^{K} \frac{F_k}{f_{max}},
\end{equation}
because all devices before the device where the last break occurs have completed their tasks. From \eqref{eq:totBroken} we see that if the local processing and transmission delay of the $(M,\mathcal{M})$-broken queue $C_{\mathrm{max \{ \mathcal{M} \} }}$ is reduced, the total inference delay of all devices will also be reduced. Driven by this observation, we will prove that by reallocating bandwidth from the device associated with the first break to the device associated with the last break, the total inference delay time is reduced. Note that such a technique is feasible because the $\left( \mathrm{min} \{ \mathcal{M} \} - 1 \right)$-th device of the initial queue can allow an increase in its delay, so that
\begin{equation}\label{eq:realloc}
     C'_{ \mathrm{min} \{ \mathcal{M} \} - 1 } 
    = C_{ \mathrm{min} \{ \mathcal{M} \} } - \frac{F_{\mathrm{min} \{ \mathcal{M}\} -1 }}{f_{\mathrm{max}}},
\end{equation}
because this ensures that $I'_{ \mathrm{min} \{ \mathcal{M} \} - 1 } = C_{ \mathrm{min} \{ \mathcal{M} \} }$, which simply eliminates the first break without changing $T_K^{(M,\mathcal{M})}$, since the latter depends only on the last break. The following lemma proves that by reallocating the bandwidth between the device corresponding to the first break and the device corresponding to the last break and using \eqref{eq:realloc}, the queuing inference delay is reduced. 
\begin{lemma}\label{lem:broken}
Let 
$M \geq 3$ and let 
$\mathcal{M}_1 = \left\{ \{ \mathcal{M},\mathrm{min} \{ \mathcal{M} \}-1 \} \backslash \{ \mathrm{max} \{ \mathcal{M} \}, \mathrm{min} \{ \mathcal{M} \} \} \right\}$ 
be the set of the intermediate $M-3$ breaks of the $(M,\mathcal{M})$-broken queue. Let also $\mathcal{M}_2 = \left\{ \{ \mathcal{M}, \mathrm{min} \{ \mathcal{M} \} \!-\! 1 , \mathrm{max} \{ \mathcal{M} \} + 1 \} \backslash \{ \mathrm{max} \{ \mathcal{M} \},  \mathrm{min} \{ \mathcal{M} \} \} \right\}$ denote the set of the last $M-2$ breaks of the $(M,\mathcal{M})$-broken queue, but with the last break occurring at the next position. Finally, let 
$\mathcal{M}_3 = \left\{ \{ \mathcal{M}, \mathrm{min} \{ \mathcal{M} \} - 1 , \mathrm{max} \{ \mathcal{M} \} + 1 \} \backslash \{  \mathrm{min} \{ \mathcal{M} \} \} \right\}$ be the set of the same breaks as the $(M,\mathcal{M})$-broken queue, but with an additional occurrence at the $\mathrm{max} \{ \mathcal{M} \} + 1$ position. All $\mathcal{M}_1, \mathcal{M}_2, \mathcal{M}_3$ contain an additional break at the $\mathrm{min} \{ \mathcal{M} \} - 1$ position. Then, the $(M,\mathcal{M})$-broken queue has a worse inference delay time than the $(M-1,\mathcal{M}_1)$-broken, $(M,\mathcal{M}_2)$-broken and $(M+1,\mathcal{M}_3)$-broken queues.
\end{lemma}
\begin{IEEEproof}
    The proof is given in Appendix A.
\end{IEEEproof}

\begin{figure}
    \centering
    \includegraphics[width=0.98\columnwidth]{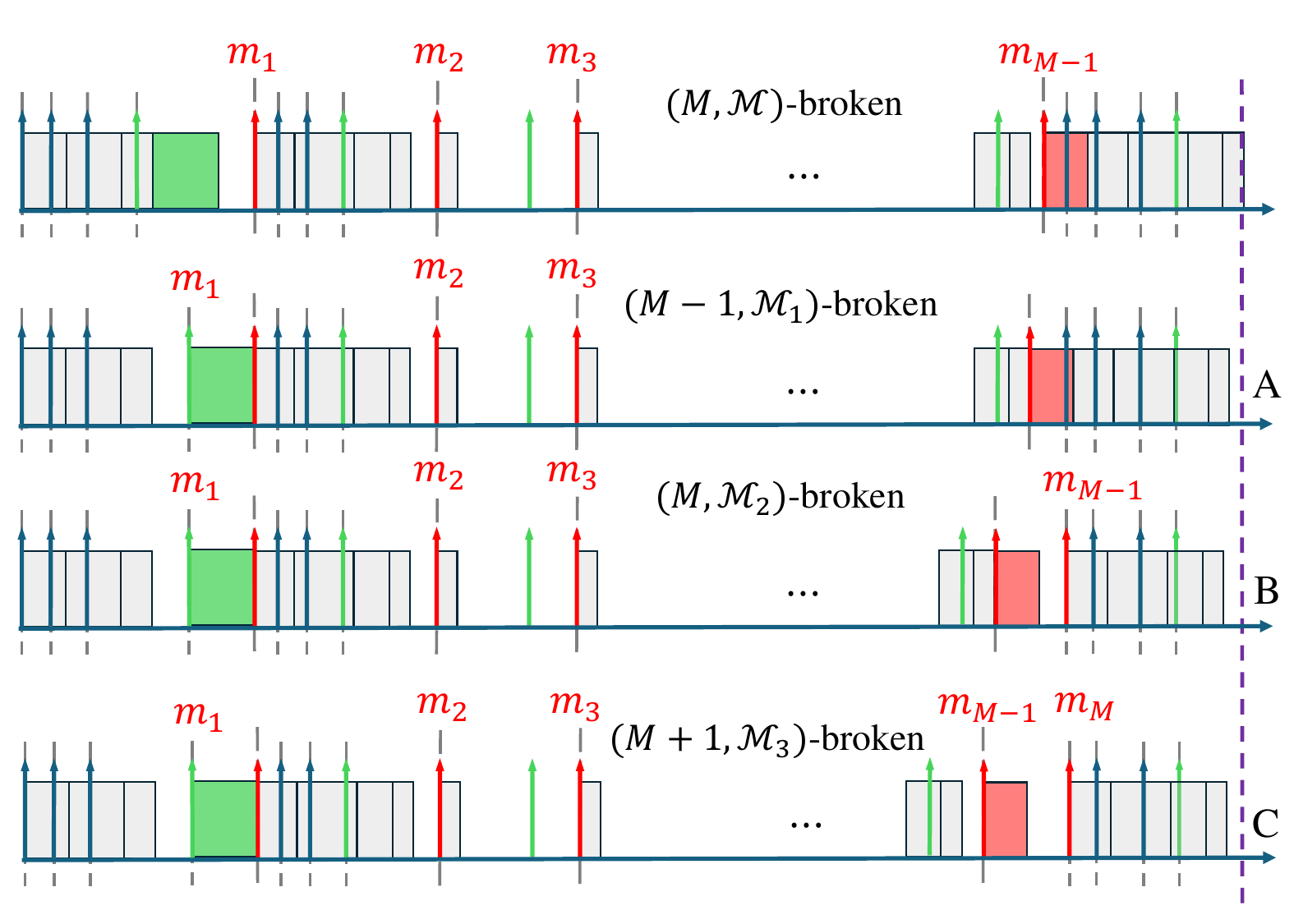}
    \caption{Different possible instances of broken queues. Arrows symbolize the order of $C_k$ and rectangles their corresponding $F_k$. In the initial queue, red arrivals indicate the first element of each unbroken sub-queue, green arrivals the last element of each unbroken sub-queue and blue arrivals the intermediate elements of each sub-queue.}
    \label{fig:queueM}
\end{figure}

From \eqref{eq:totBroken} and Lemma \ref{lem:broken}, we can conclude that reducing the local processing and transmission delay of the ``$M$-th unbroken queue, as shown in Fig. \ref{fig:queueM}, reduces the overall inference delay for all devices. This is because when the maximum delay \( C_{\mathrm{max \{ \mathcal{M} \} }} \) is reduced, one of three outcomes occurs, each of which leads to improved performance. First, the last break in the queue can be eliminated without creating a new one, resulting in an $(M-1,\mathcal{M}_1)$-broken scenario shown as A in Fig. \ref{fig:queueM}. Alternatively, the last break could be eliminated and a new one created at the next position, resulting in an $(M,\mathcal{M}_2)$-broken scenario described as B in Fig. \ref{fig:queueM}. Finally, the last break could be preserved, with a new break appearing at the next position, resulting in an $(M+1,\mathcal{M}_3)$-broken scenario, which is given as C in Fig. \ref{fig:queueM}. In all cases, as shown by Lemma \ref{lem:broken}, each of these outcomes performs better than the original configuration.

Therefore, we propose a heuristic algorithm that can improve the total inference delay time by reallocating bandwidth from the devices of the first break to those of the last break, with the goal of reducing $C_{\mathrm{max \{ \mathcal{M} \} }}$. Using \eqref{eq:Constants}, \eqref{eq:ConstantsF} and \eqref{eq:realloc} we can calculate the reallocated bandwidth for the $\left( \mathrm{min \{ \mathcal{M} \} - 1} \right)$-th device of the initial queue by solving the resulting equation in terms of the new bandwidth $B'_{\mathrm{min \{ \mathcal{M} \} - 1}}$, which is illustrated in Fig. \ref{fig:queueM} as a shift of the first green arrival time and its rectangle. Then the excess bandwidth $\Delta B = B_{\mathrm{min \{ \mathcal{M} \} - 1}} - B'_{\mathrm{min \{ \mathcal{M} \} - 1}}$ can be reassigned to the $\mathrm{max \{ \mathcal{M} \}}$-th device in the queue, which has an increased bandwidth of $ B'_{\mathrm{max \{ \mathcal{M} \} }} = B_{\mathrm{max \{ \mathcal{M} \} }} + \Delta B $. By iteratively following the same procedure until all but one break is eliminated, we can reduce the total inference delay time as described in Algorithm \ref{alg:heuristicAlgo}.

\begin{algorithm}\label{alg:heuristicAlgo}
\caption{Alternating Optimization implementing Heuristic Algorithm for Serial Processing}
Fix number of iterations, \textit{iters}. \\

For each device perform the \textit{slicing policy} $a_{k,l},\ \forall k \in \mathcal{K}$. \\

\For{$j = 1:\textit{iters}$}{
    Fix bandwidth for each device equal to $B_k = B_{\mathrm{tot}}/K$. \\
    
    Sort the devices in terms of $C_{k}$ and calculate $I_k$, characterizing the queue.

    \While{$|\mathcal{M}| > 2$}{        Find the number of ``breaks" and create $\mathcal{M}$. \\
    
        Solve \eqref{eq:realloc} to get $B'_{\mathrm{min \{ \mathcal{M} \} - 1}}$. \\
    
        Calculate bandwidth excess, $\Delta B = B_{\mathrm{min \{ \mathcal{M} \} - 1}} - B'_{\mathrm{min \{ \mathcal{M} \} - 1}}$. \\
    
        Change bandwidth of the $\mathrm{max \{ \mathcal{M} \}}$-th element of the queue to $ B'_{\mathrm{max \{ \mathcal{M} \} }} =  B_{\mathrm{max \{ \mathcal{M} \} }} + \Delta B $. \\
    }
    
    Find optimal \textit{slicing policy} $a_{k,l},\ \forall k \in \mathcal{K}$ as in \eqref{eq:minmaxDevice}. \\
}

Keep minimum achieving \textit{slicing policy} $a_{k,l},\ \forall k \in \mathcal{K}$ and bandwidth allocation vector. \\
\end{algorithm}

\subsection{Complexity Analysis}
In this subsection, we compare the proposed schemes in terms of complexity. For the parallel processing schemes, \eqref{eq:OptProblemSimple} can significantly reduce the complexity compared to \eqref{eq:OptProblemGeneral}. Unlike many convex optimization techniques, the proposed one does not need more iterations, since only the first root of the equation in \eqref{eq:equationClosed} is computed, which is in a small interval. If \eqref{eq:OptProblemSimple} is solved with \eqref{eq:equationClosed}, it is easy to see that the complexity is given as $\mathcal{O}(2|\mathcal{F}| ) \approx \mathcal{O}(2K)$, since only the minimum of the two sequences described by \eqref{eq:conditions} is required. On the other hand, \eqref{eq:OptProblemGeneral} may perform better, but it has twice as many variables and its complexity can be derived as $\mathcal{O}\left( (2K)^{3.5} \right)$ \cite{anstreicher1999linear}, showing a large gap between the two that needs to be considered for practical deployment designs.

Regarding the complexity of the serial processing techniques, the convex optimization problem in \eqref{eq:minmaxDevice} has a complexity of \(\mathcal{O}\left( K^{3.5} \right)\), as established in \cite{anstreicher1999linear}. In contrast, the proposed heuristic algorithm (lines 4–10 in Algorithm \ref{alg:heuristicAlgo}) requires sorting the estimated arrival times \(C_k\), with a maximum of \(K\) iterations if there are \(K\) breaks. Additionally, it solves up to \(K\) single-variable equations, resulting in a much lower complexity than \(\mathcal{O}\left( K^{3.5} \right)\).
Notably, in each iteration, the updated set \(\mathcal{M}\) only varies by 1–2 elements in known positions (transforming into \(\mathcal{M}_1\), \(\mathcal{M}_2\), or \(\mathcal{M}_3\)), which eliminates the need for a complete re-evaluation of \(\mathcal{M}\). Therefore, in the worst case, the overall complexity of Algorithm \ref{alg:heuristicAlgo} is \(\mathcal{O}(K \log_2(K) + K)\).
For completeness, lines $1-4$ in Table \ref{tab:complexities} summarize the overall complexity of each scheme considered, including the complexity of the SL-inspired slicing policy, which adds an additional \(\mathcal{O}(KL)\) term over all schemes because of the requirement to identify the minimum of each \(C_{k,l}\),~\(\forall \in \{1, \cdots, L\}\) for all $K$ devices.

\begin{table}
    \centering
    \caption{Complexity Comparison.}
    \begin{tabular}{c||c}
    \hline \hline
    Proposed \eqref{eq:OptProblemGeneral} & $\mathcal{O}\left( (2K)^{3.5}  + KL \right)$  \\
    \hline
    Proposed \eqref{eq:OptProblemSimple} & $\mathcal{O}\left( 2K  + KL \right)$  \\
    \hline 
    Queue \eqref{eq:minmaxDevice} & $\mathcal{O}\left(K^{3.5}  + KL \right)$  \\
    \hline
    Queue heuristic (Alg. \ref{alg:heuristicAlgo}) & $\mathcal{O}\left(K\log_2(K) + K + KL \right)$  \\
    \hline
    Min-data layer & $\mathcal{O}\left( (2K)^{3.5} \right)$  \\
    \hline
    First layer & $\mathcal{O}\left( (2K)^{3.5} \right)$  \\
    \hline
    Queue first layer & $\mathcal{O}\left( K^{3.5} \right)$  \\
    \hline \hline
    \end{tabular}
    \label{tab:complexities}
\end{table}




\section{Numerical Results and Discussion}\label{sec:num}
In this section, numerical results illustrating the performance of all considered schemes are presented. To study the proposed policy, ENet was used as the semantic segmentation architecture at each device in the network, although different architectures can also be utilized across the devices. Moreover, to evaluate the inference delay performance, high-quality RGB images with dimensions of $1024 \times 2048$ are processed, which is typical for many CV applications. Similar simulations can be performed for other architectures and different dimensions, depending on the application of interest. The simulation parameters are given in Table \ref{tab:ParamsNumerical} and for all figures Monte Carlo simulations over $500$ different realizations have been performed under Rayleigh channel fading conditions, i.e., $h_k \sim CN(0,1)$. 

\begin{table}
    \centering
    \caption{Simulation parameters.}
    \begin{tabular}{c||c}
    \hline \hline
    Initial Data (bits),  $D_{k,1}$ &$192$ Mbit  \\
    \hline
    Final channel outputs,  $C_{out}$ & 20 \\
    \hline 
    Wavelength, $\lambda$ & $0.05$ m \\
    \hline
    Distance from BS, $d$ & $50$ m \\
    \hline
    Path loss exponent, $n$ & $2.4$ \\
    \hline
    Transmit power, $P_{k}$ & $1$ W \\
    \hline
    Antenna gains, $G_t , G_r$ & $1 , 10$ dBi \\
    \hline
    Total bandwidth, $B_{\mathrm{tot}}$ & $200$ MHz \\
    \hline
    Number of devices, $K$ & $10$ \\
    \hline
    Number of BMs in ENet, $L$ & $30$ \\
    \hline
    Device computational resources, $f_k$ & $30$ GFLOPS \\
    \hline
    Central server computational resources, $f_{\mathrm{max}}$ & $300$ GFLOPS \\
    \hline \hline
    \end{tabular}
    \label{tab:ParamsNumerical}
\end{table}

The considered schemes for the case of \emph{parallel} data processing at the central server are the following:
\begin{itemize}
    \item Proposed \eqref{eq:OptProblemGeneral}: This is the optimal policy and results from solving the optimization problem \eqref{eq:OptProblemGeneral}, which aims to maximize performance at the expense of increased complexity.
    \item Proposed \eqref{eq:OptProblemSimple}: This is a suboptimal policy derived from the solution of \eqref{eq:OptProblemSimple}, thus having a lower complexity than the previous policy.
    \item Min-data layer: Each device chooses to implement the \textit{slicing policy} at the first BM corresponding to the minimum amount of data to be transmitted. Its complexity is lower than solving \eqref{eq:OptProblemGeneral}, but greater than solving \eqref{eq:OptProblemSimple}.
    \item First layer: This is essentially a no split-learning policy, which is used to demonstrate the benefits of SL for real-time semantic fragmentation. Its complexity is equal to that of the ``min-data layer" policy.
\end{itemize}
Similarly, the investigated schemes for \emph{serial} processing are:
\begin{itemize}
    \item Queue \eqref{eq:minmaxDevice}: This policy results from the solution of \eqref{eq:minmaxDevice}, and it has the greatest complexity among all ``queuing" schemes. 
    \item Queue heuristic: This policy is derived from the iterative algorithm \ref{alg:heuristicAlgo} and aims to maximize performance at the expense of reduced complexity, compared to the previous scheme.
    \item Queue first layer: This policy is similar to the ``first layer" policy.
\end{itemize}
The complexity of each scheme is given in Table \ref{tab:complexities}. Note that the schemes that utilize a fixed \textit{slicing policy} do not have the extra term $\mathcal{O}(KL)$ because they do not have a selection option.

\begin{figure}
    \centering
    \begin{tikzpicture}
        \begin{axis}[
            width=0.95\linewidth,
            xlabel = {Number of devices $K$},
            ylabel = {Average maximum delay (s)},
            ymin = 0.8,
            ymax = 3.8,
            xmin = 4,
            xmax = 16,
            ytick = {0.8,1.3,...,3.8},
            scaled y ticks=false, 
            grid = both,
            minor grid style={gray!25},
            major grid style={gray!50},
            legend columns=1, 
		legend entries ={Proposed {\eqref{eq:OptProblemGeneral}},{Proposed \eqref{eq:OptProblemSimple}},{Min-data layer},{First layer},{Queue heuristic},{Queue \eqref{eq:minmaxDevice}}, {Queue first layer}},
            legend cell align = {left},
            legend style={font=\footnotesize},
            legend style={at={(0,1)},anchor=north west},
            legend image post style={scale=0.7}, 
            ]
            \addplot[
            black,
            mark = square,
            mark repeat = 1,
            mark size = 2,
            line width = 1pt,
            style = solid,
            ]
            table {NewDevices/full_optimal_dev.dat};
            \addplot[
            red,
            mark = o,
            mark repeat = 1,
            mark size = 2,
            line width = 1pt,
            style = solid,
            ]
            table {NewDevices/equation_optimal_dev.dat};
            \addplot[
            blue,
            mark = asterisk,
            mark repeat = 1,
            mark size = 3,
            line width = 1pt,
            style = solid,
            ]
            table {NewDevices/mid_dev.dat};
            \addplot[
            green,
            mark = diamond,
            mark repeat = 1,
            mark size = 3,
            line width = 1pt,
            style = solid,
            ]
            table {NewDevices/direct_dev.dat};
            \addplot[
            gray,
            mark = x,
            mark repeat = 1,
            mark size = 3,
            line width = 1pt,
            style = solid,
            ]
            table {Queue/NewDevices/algo_optimal_dev.dat};
            \addplot[
            magenta,
            mark = triangle,
            mark repeat = 1,
            mark size = 2,
            line width = 1pt,
            style = solid,
            ]
            table {Queue/NewDevices/con_optimal_dev.dat};
            \addplot[
            cyan,
            mark = star,
            mark repeat = 1,
            mark size = 3,
            line width = 1pt,
            style = solid,
            ]
            table {Queue/NewDevices/direct_optimal_dev.dat};
        \end{axis}
    \end{tikzpicture}
    \caption{Average maximum delay versus number of devices.}
    \label{fig:devices_vary}
\end{figure}
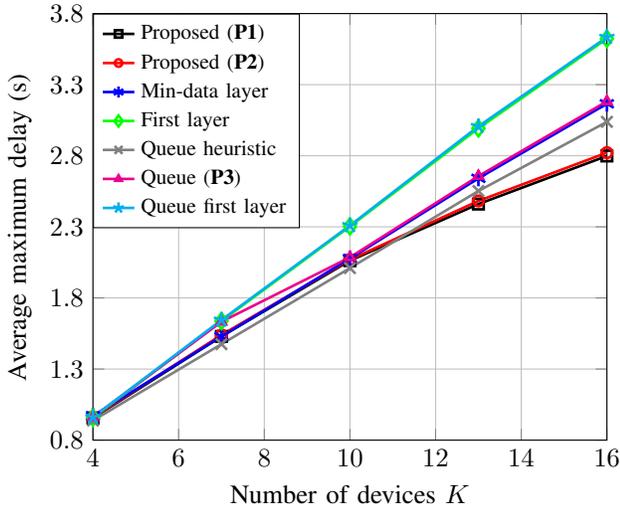

Fig. \ref{fig:devices_vary} illustrates the relationship between the average maximum delay and the number of devices \( K \) for different SL policies. First, it is evident that both ``first layer'' policies perform worse than any SL-based policy, underscoring the effectiveness of SL in reducing inference delay. Notably, there is no single optimal policy for all scenarios. When there are fewer devices, the ``queue heuristic'' policy yields the best performance, as it fully utilizes the server's resources per device while maintaining minimal queuing delay. However, as the number of devices increases, the delay for this scheme increases linearly, resulting in reduced efficiency compared to the proposed \eqref{eq:OptProblemGeneral} and \eqref{eq:OptProblemSimple} policies. Among them, the ``proposed \eqref{eq:OptProblemSimple}'' policy achieves the same performance as the more complex ``proposed \eqref{eq:OptProblemGeneral}'' policy, but with significantly reduced computational complexity, making it preferable for real-time applications. In addition, the ``min-data layer'' policy initially matches the performance of \eqref{eq:OptProblemGeneral}, indicating that the optimal slicing layer is initially the layer corresponding to the least data to be transmitted. However, as the number of users increases, the optimal slicing layer shifts away from the one with the least amount of data transmitted, because this layer causes tasks to arrive at the same time, increasing the workload on the server.

\begin{figure}
    \centering
    \begin{tikzpicture}
        \begin{axis}[
            width=0.95\linewidth,
            xlabel = {Transmit power $P_k$ (W)},
            ylabel = {Average maximum delay (s)},
            ymin = 1.8,
            ymax = 2.6,
            xmin = 0.2,
            xmax = 2,
            xtick = {0.2,0.5,...,2},
            scaled y ticks=false, 
            grid = both,
            minor grid style={gray!25},
            major grid style={gray!50},
            legend columns=2, 
		legend entries ={Proposed {\eqref{eq:OptProblemGeneral}},{Proposed \eqref{eq:OptProblemSimple}},{Min-data layer},{First layer},{Queue heuristic},{Queue \eqref{eq:minmaxDevice}}, {Queue first layer}},
            legend cell align = {left},
            legend style={font=\footnotesize},
            legend style={at={(1,1)},anchor=north east},
            legend image post style={scale=0.5}, 
            ]
            \addplot[
            black,
            mark = square,
            mark repeat = 1,
            mark size = 2,
            line width = 1pt,
            style = solid,
            ]
            table {NewPower/full_optimal_p.dat};
            \addplot[
            red,
            mark = o,
            mark repeat = 1,
            mark size = 2,
            line width = 1pt,
            style = solid,
            ]
            table {NewPower/equation_optimal_p.dat};
            \addplot[
            blue,
            mark = asterisk,
            mark repeat = 1,
            mark size = 3,
            line width = 1pt,
            style = solid,
            ]
            table {NewPower/mid_p.dat};
            \addplot[
            green,
            mark = diamond,
            mark repeat = 1,
            mark size = 3,
            line width = 1pt,
            style = solid,
            ]
            table {NewPower/direct_p.dat};
            \addplot[
            gray,
            mark = x,
            mark repeat = 1,
            mark size = 3,
            line width = 1pt,
            style = solid,
            ]
            table {Queue/NewPower/algo_optimal_p.dat};
            \addplot[
            magenta,
            mark = triangle,
            mark repeat = 1,
            mark size = 2,
            line width = 1pt,
            style = solid,
            ]
            table {Queue/NewPower/con_optimal_p.dat};
            \addplot[
            cyan,
            mark = star,
            mark repeat = 1,
            mark size = 3,
            line width = 1pt,
            style = solid,
            ]
            table {Queue/NewPower/direct_optimal_p.dat};
        \end{axis}
    \end{tikzpicture}
    \caption{Average maximum delay versus transmit power.}
    \label{fig:power_vary}
\end{figure}
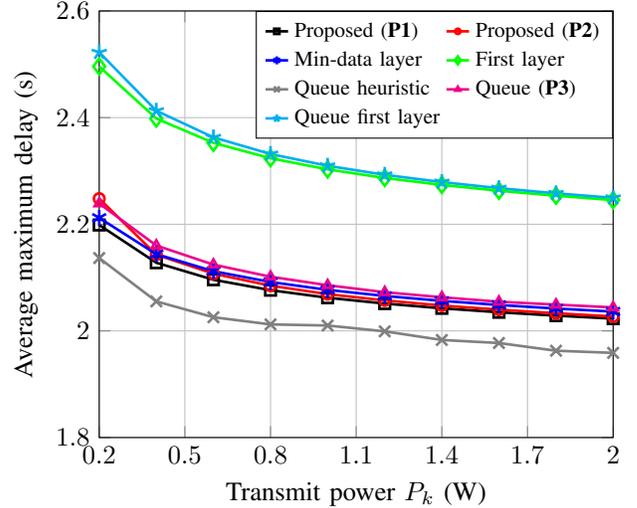

Fig. \ref{fig:power_vary} illustrates the relationship between transmit power \( P_k \) and average maximum delay. As transmit power increases, all policies show a reduction in delay, highlighting the benefit of higher transmit rates in reducing communication time. Notably, the effect of transmit power on latency is consistent across all schemes. Given that this figure represents a ten-device setup, the queue heuristic emerges as the most effective scheme across all power levels, consistent with the findings in Fig. \ref{fig:devices_vary}. In addition, the ``first layer'' policies yield the highest delays, even with increasing transmit power, indicating that improving communication quality alone is not sufficient to achieve real-time CV performance. 


Fig. \ref{fig:bandwidth_vary} illustrates the impact of available bandwidth on delay. Notably, the ``proposed \eqref{eq:OptProblemGeneral}" and ``proposed \eqref{eq:OptProblemSimple}'' policies show nearly identical performance, even though the former uses optimal bandwidth allocation but not optimal resource allocation at the server. This similarity in performance is due to the small number of devices, which results in an almost equal distribution of bandwidth. Allocating bandwidth evenly across devices avoids differences in task arrival times and reduces the maximum inference delay, which is consistent with the performance of the ``proposed \eqref{eq:OptProblemSimple}'' policy.
It is also important to note that while the ``queue \eqref{eq:minmaxDevice}'' policy initially aligns with the parallel processing policies, it shifts toward the ``first tier'' policies as bandwidth increases. This shift occurs because as bandwidth increases, the transmission delay becomes much smaller compared to the local processing delay of each slicing policy used.  Similar to the effect of transmit power, increasing bandwidth uniformly reduces delay across all schemes, although the improvements taper off as bandwidth increases. This again demonstrates that physical layer enhancements alone are not sufficient to achieve real-time performance for CV applications.

\begin{figure}
    \centering
    \begin{tikzpicture}
        \begin{axis}[
            width=0.95\linewidth,
            xlabel = {Total available bandwidth $B_{\mathrm{tot}}$ (MHz)},
            ylabel = {Average maximum delay (s)},
            ymin = 1.6,
            ymax = 2.6,
            xmin = 150,
            xmax = 400,
            xtick = {150,200,250,300,350,400},
            xticklabels={150,200,250,300,350,400},
            scaled y ticks=false, 
            grid = both,
            minor grid style={gray!25},
            major grid style={gray!50},
            legend columns=2, 
		legend entries ={Proposed {\eqref{eq:OptProblemGeneral}},{Proposed \eqref{eq:OptProblemSimple}},{Min-data layer},{First layer},{Queue heuristic},{Queue \eqref{eq:minmaxDevice}}, {Queue first layer}},
            legend cell align = {left},
            legend style={font=\footnotesize},
            legend style={at={(1,1)},anchor=north east},
            legend image post style={scale=0.5}, 
            ]
            \addplot[
            black,
            mark = square,
            mark repeat = 1,
            mark size = 2,
            line width = 1pt,
            style = solid,
            ]
            table {NewBandwidth/full_optimal_b.dat};
            \addplot[
            red,
            mark = o,
            mark repeat = 1,
            mark size = 2,
            line width = 1pt,
            style = solid,
            ]
            table {NewBandwidth/equation_optimal_b.dat};
            \addplot[
            blue,
            mark = asterisk,
            mark repeat = 1,
            mark size = 3,
            line width = 1pt,
            style = solid,
            ]
            table {NewBandwidth/mid_b.dat};
            \addplot[
            green,
            mark = diamond,
            mark repeat = 1,
            mark size = 3,
            line width = 1pt,
            style = solid,
            ]
            table {NewBandwidth/direct_b.dat};
            \addplot[
            gray,
            mark = x,
            mark repeat = 1,
            mark size = 3,
            line width = 1pt,
            style = solid,
            ]
            table {Queue/NewBandwidth/algo_optimal_b.dat};
            \addplot[
            magenta,
            mark = triangle,
            mark repeat = 1,
            mark size = 2,
            line width = 1pt,
            style = solid,
            ]
            table {Queue/NewBandwidth/con_optimal_b.dat};
            \addplot[
            cyan,
            mark = star,
            mark repeat = 1,
            mark size = 3,
            line width = 1pt,
            style = solid,
            ]
            table {Queue/NewBandwidth/direct_optimal_b.dat};
        \end{axis}
    \end{tikzpicture}
    \caption{Average maximum delay versus total available bandwidth.}
    \label{fig:bandwidth_vary}
\end{figure}
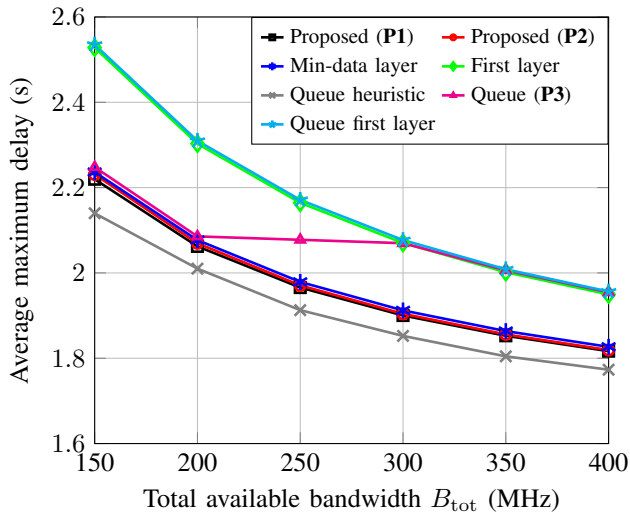

\begin{figure}
    \centering
    \begin{tikzpicture}
        \begin{axis}[
            width=0.95\linewidth,
            xlabel = {Device frequency resources $f_k$ (GFLOPS)},
            ylabel = {Average maximum delay (s)},
            ymin = 1.8,
            ymax = 2.6,
            xmin = 10,
            xmax = 40,
            xtick = {10,15,20,...,40},
            scaled y ticks=false, 
            grid = both,
            minor grid style={gray!25},
            major grid style={gray!50},
            legend columns=2, 
		legend entries ={Proposed {\eqref{eq:OptProblemGeneral}},{Proposed \eqref{eq:OptProblemSimple}},{Min-data layer},{First layer},{Queue heuristic},{Queue \eqref{eq:minmaxDevice}}, {Queue first layer}},
            legend cell align = {left},
            legend style={font=\footnotesize},
            legend style={at={(1,1)},anchor=north east},
            legend image post style={scale=0.7}, 
            ]
            \addplot[
            black,
            mark = square,
            mark repeat = 1,
            mark size = 2,
            line width = 1pt,
            style = solid,
            ]
            table {NewF_vehicles/full_optimal_fveh.dat};
            \addplot[
            red,
            mark = o,
            mark repeat = 1,
            mark size = 2,
            line width = 1pt,
            style = solid,
            ]
            table {NewF_vehicles/equation_optimal_fveh.dat};
            \addplot[
            blue,
            mark = asterisk,
            mark repeat = 1,
            mark size = 3,
            line width = 1pt,
            style = solid,
            ]
            table {NewF_vehicles/mid_fveh.dat};
            \addplot[
            green,
            mark = diamond,
            mark repeat = 1,
            mark size = 3,
            line width = 1pt,
            style = solid,
            ]
            table {NewF_vehicles/direct_fveh.dat};
            \addplot[
            gray,
            mark = x,
            mark repeat = 1,
            mark size = 3,
            line width = 1pt,
            style = solid,
            ]
            table {Queue/NewF_vehicles/algo_optimal_fveh.dat};
            \addplot[
            magenta,
            mark = triangle,
            mark repeat = 1,
            mark size = 2,
            line width = 1pt,
            style = solid,
            ]
            table {Queue/NewF_vehicles/con_optimal_fveh.dat};
            \addplot[
            cyan,
            mark = star,
            mark repeat = 1,
            mark size = 3,
            line width = 1pt,
            style = solid,
            ]
            table {Queue/NewF_vehicles/direct_optimal_fveh.dat};
        \end{axis}
    \end{tikzpicture}
    \caption{Average maximum delay versus device frequency resources.}
    \label{fig:fveh_vary}
\end{figure}
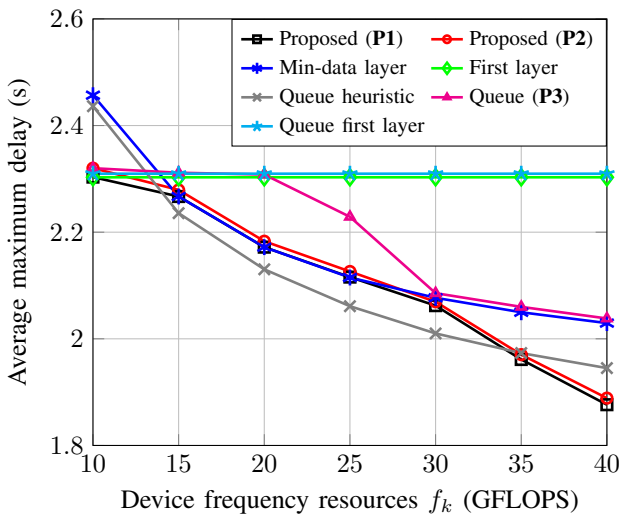


Fig. \ref{fig:fveh_vary} shows the effect of device frequency resources $f_k$ on the average maximum delay for different split learning policies in a distributed semantic segmentation framework. As illustrated when the devices' resources are small immediate transmission is optimal because the transmission delay is less than the local processing time, which is also the reason that the ``min-data layer" policy is outperformed. For mid-range device computational resources a shift towards the ``min-data layer" policy is observed for the parallel processing schemes. In this range we also notice that ``queue heuristic" is the optimal scheme, which is a result of the reduced remaining processing at the server, $F_k$, causing more breaks to exist between arrival times leading to utilization of Algorithm \ref{alg:heuristicAlgo}. In addition, for large device resources the ``proposed \eqref{eq:OptProblemGeneral}" and ``proposed \eqref{eq:OptProblemSimple}" schemes outperform ``queue heuristic" because the parallel processing schemes allow a more flexible \textit{slicing policy} realization. The latter also implements the \textit{slicing policy}, but it appears to be less flexible due to the fact that queuing at the server can act as a bottleneck compared to parallel processing when $F_k$ at the server side is quite small due to increased $f_k$ at the devices. 

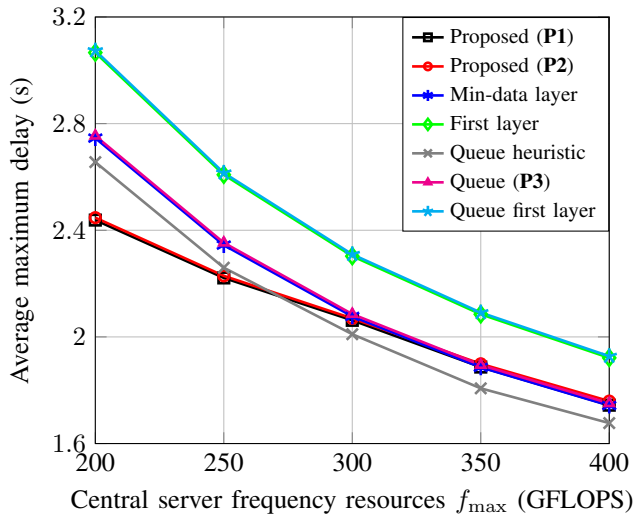
\begin{figure}
    \centering
    \begin{tikzpicture}
        \begin{axis}[
            width=0.95\linewidth,
            xlabel = {Central server frequency resources $f_{\mathrm{max}}$ (GFLOPS)},
            ylabel = {Average maximum delay (s)},
            ymin = 1.6,
            ymax = 3.2,
            xmin = 200,
            xmax = 400,
            xtick = {200,250,300,350,400},
            xticklabels={200,250,300,350,400},
            ytick = {1.6,2,...,3.2},
            scaled y ticks=false, 
            grid = both,
            minor grid style={gray!25},
            major grid style={gray!50},
            legend columns=1, 
		legend entries ={Proposed {\eqref{eq:OptProblemGeneral}},{Proposed \eqref{eq:OptProblemSimple}},{Min-data layer},{First layer},{Queue heuristic},{Queue \eqref{eq:minmaxDevice}}, {Queue first layer}},
            legend cell align = {left},
            legend style={font=\footnotesize},
            legend style={at={(1,1)},anchor=north east},
            legend image post style={scale=0.7}, 
            ]
            \addplot[
            black,
            mark = square,
            mark repeat = 1,
            mark size = 2,
            line width = 1pt,
            style = solid,
            ]
            table {NewF_server/full_optimal_fser.dat};
            \addplot[
            red,
            mark = o,
            mark repeat = 1,
            mark size = 2,
            line width = 1pt,
            style = solid,
            ]
            table {NewF_server/equation_optimal_fser.dat};
            \addplot[
            blue,
            mark = asterisk,
            mark repeat = 1,
            mark size = 3,
            line width = 1pt,
            style = solid,
            ]
            table {NewF_server/mid_fser.dat};
            \addplot[
            green,
            mark = diamond,
            mark repeat = 1,
            mark size = 3,
            line width = 1pt,
            style = solid,
            ]
            table {NewF_server/direct_fser.dat};
            \addplot[
            gray,
            mark = x,
            mark repeat = 1,
            mark size = 3,
            line width = 1pt,
            style = solid,
            ]
            table {Queue/NewF_server/algo_optimal_fser.dat};
            \addplot[
            magenta,
            mark = triangle,
            mark repeat = 1,
            mark size = 2,
            line width = 1pt,
            style = solid,
            ]
            table {Queue/NewF_server/con_optimal_fser.dat};
            \addplot[
            cyan,
            mark = star,
            mark repeat = 1,
            mark size = 3,
            line width = 1pt,
            style = solid,
            ]
            table {Queue/NewF_server/direct_optimal_fser.dat};
        \end{axis}
    \end{tikzpicture}
    \caption{Average maximum delay versus central server frequency resources.}
    \label{fig:fser_vary}
\end{figure}

Fig. \ref{fig:fser_vary} shows the relationship between the central server frequency resources $f_{\max}$ and the total average maximum delay for different shared learning policies. For all policies, increasing $f_{\max}$ leads to a reduction in the average maximum delay, highlighting the benefit of increased processing capacity at the BS for minimizing inference latency. As in the previous figures the ``first layer" and ``queue first layer" policies have almost identical performance and achieve large delays compared to other policies, which is expected since both policies don't take advantage of the available devices' resources. As for the rest of the parallel processing policies, we observe that all three converge as server resources increase. This can be explained by the fact that as the resources increase, the arrival times $C_k$ become dominant over the remaining $F_k/f_{k,s}$, which are minimal. Therefore, by \eqref{eq:OptProblemGeneral} all devices will aim to have identical $C_k$, which is also the goal of \eqref{eq:minmaxDevice} expressed by the ``queue \eqref{eq:minmaxDevice}'' policy, which also converges with the other parallel schemes. However, for lower server resources, there is a large gap between the ``min-data layer'' and the ``proposed \eqref{eq:OptProblemGeneral}'' and ``proposed \eqref{eq:OptProblemSimple}'' policies, which can be attributed to the freedom of splitting due to the implemented \textit{slicing policy} of the latter two schemes. Note also the behavior of the ``queue heuristic'' policy, which shows better performance as $f_{\mathrm{max}}$ increases. This can be explained by the fact that as the resources increase, the corresponding processing time of the remaining workload at the server $F_k$ decreases, thus creating breaks between the arriving data, which will lead to the Algorithm \ref{alg:heuristicAlgo} to be used extensively. This provides improved performance over its other serial and parallel processing counterparts.

\begin{figure}
    \centering
    \begin{tikzpicture}
        \begin{axis}[
            width=0.95\linewidth,
            xlabel = {Number of iterations},
            ylabel = {Average maximum delay (s)},
            ymin = 2,
            ymax = 3.6,
            xmin = 1,
            xmax = 4,
            xtick = {1,2,...,4},
            ytick = {2,2.4,2.8,3.2,3.6},
            scaled y ticks=false, 
            grid = both,
            minor grid style={gray!25},
            major grid style={gray!50},
            legend columns=1, 
		legend entries ={Proposed {\eqref{eq:OptProblemGeneral}},{Proposed \eqref{eq:OptProblemSimple}},{Queue heuristic},{Queue \eqref{eq:minmaxDevice}}},
            legend cell align = {left},
            legend style={font=\footnotesize},
            legend style={at={(1,1)},anchor=north east},
            legend image post style={scale=0.7}, 
            ]
            \addplot[
            black,
            mark = square,
            mark repeat = 1,
            mark size = 2,
            line width = 1pt,
            style = solid,
            ]
            table {Iters/iter_parallel_full.dat};
            \addplot[
            red,
            mark = o,
            mark repeat = 1,
            mark size = 2,
            line width = 1pt,
            style = solid,
            ]
            table {Iters/iter_parallel_equation.dat};
            \addplot[
            gray,
            mark = x,
            mark repeat = 1,
            mark size = 3,
            line width = 1pt,
            style = solid,
            ]
            table {Iters/iter_serial_algo.dat};
            \addplot[
            magenta,
            mark = triangle,
            mark repeat = 1,
            mark size = 2,
            line width = 1pt,
            style = solid,
            ]
            table {Iters/iter_serial_con.dat};
        \end{axis}
    \end{tikzpicture}
    \caption{Average maximum delay versus number of iterations.}
    \label{fig:iters_vary}
\end{figure}
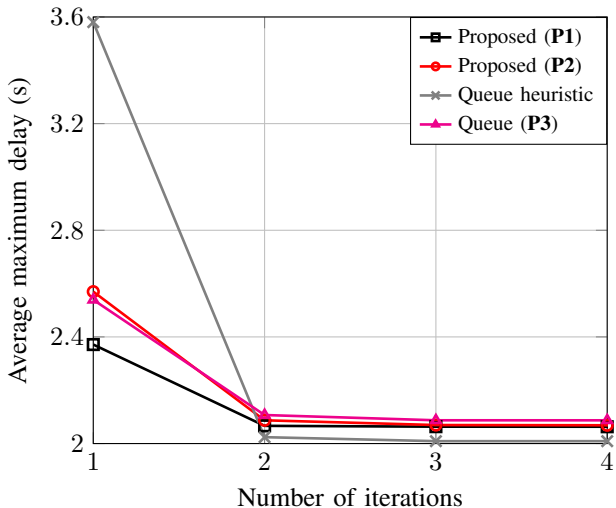

Fig. \ref{fig:iters_vary} illustrates the effect of iterations in the policies that use alternating optimization. As shown in the first iteration, the more complex schemes perform significantly better than their less complex counterparts, as indicated by the pairs ``proposed \eqref{eq:OptProblemGeneral}" - ``proposed \eqref{eq:OptProblemSimple}" and ``queue \eqref{eq:minmaxDevice}" - ``queue heuristic", respectively. However, all proposed schemes converge to their final solution within a minimal number of iterations, with a marginal improvement by increasing the number of iterations above $3$. This means that the complexity of these schemes does not increase with the number of iterations, and according to Table \ref{tab:complexities}, the overall complexity of each scheme can be calculated by multiplying it by the number of iterations, which does not affect it much. It is significant to emphasize that these alternating schemes are also little affected by the utilization of the \textit{slicing policy} in terms of complexity, since the latter only adds the linear term $\mathcal{O}(KL)$, highlighting another advantage of it. Moreover, the fast convergence shows that the utilization of the \textit{slicing policy} not only improves performance, but also allows the use of low-complexity techniques with identical or even better performance than their high-complexity counterparts, which is critical especially for real-time CV applications.

\section{Conclusions} \label{sec:concl}
In this work, we introduced a novel approach to minimize inference delay in semantic segmentation by leveraging SL to meet the requirements of real-time CV applications on resource-constrained devices. Recognizing the limitations of traditional centralized processing methods, we applied SL to partition DNNs between edge devices and a central server. This partitioning enabled localized data processing, significantly reduced data transmission requirements, and decreased inference delays. Our approach involved the joint optimization of bandwidth allocation, cut layer selection within DNNs on edge devices, and resource allocation on the central server. By studying both parallel and serial data processing scenarios, we developed and validated low-complexity heuristic solutions that closely approximated optimal performance while reducing computational overhead. Numerical results showed that our SL-based approach effectively minimized inference delay, highlighting its potential to improve the responsiveness and efficiency of real-time CV applications in distributed, resource-constrained environments. This research lays the foundation for future studies on scalable SL methods and further optimization of distributed CV applications.

\appendices
\section{Proof of Lemma 2}
Starting from \eqref{eq:totBroken}, we can equivalently write 
\begin{align}\label{eq:infInequalityM1}
T_K^{(M,\mathcal{M})} &= C_{\mathrm{max \{ \mathcal{M} \} }} + \sum\limits_{k = \mathrm{max \{ \mathcal{M} \} }}^{K} \frac{F_k}{f_{\mathrm{max}}} \nonumber \\
      &> I_{\mathrm{max \{ \mathcal{M} \} - 1}} + \sum\limits_{k = \mathrm{max \{ \mathcal{M} \} }}^{K} \frac{F_k}{f_{\mathrm{max}}} \nonumber \\
      &= C_{\mathrm{max} \{ \mathcal{M}_1 \}} + \sum\limits_{k = \mathrm{max} \{ \mathcal{M}_1 \}}^{K} \frac{F_k}{f_{\mathrm{max}}} \nonumber \\
      &= T_K^{(M,\mathcal{M}_1)} = T_K^{(M-1,\mathcal{M}_1)},
\end{align}
where the third step is a consequence of \eqref{eq:unbrokenQ} in its sub-queue and the last equality holds by the definition of the inference delay time of the corresponding queue and the fact that the latter does not depend on the number of breaks but on the position of the last break. It also holds that 
\begin{align}\label{eq:infInequalityM}
T_K^{(M,\mathcal{M})} &= C_{\mathrm{max \{ \mathcal{M} \} }} + \sum\limits_{k = \mathrm{max \{ \mathcal{M} \} }}^{K} \frac{F_k}{f_{\mathrm{max}}} \nonumber \\
&= I_{\mathrm{max \{ \mathcal{M} \} }} + \sum\limits_{k = \mathrm{max \{ \mathcal{M} \} + 1 }}^{K} \frac{F_k}{f_{\mathrm{max}}} \nonumber \\
      &> C_{\mathrm{max \{ \mathcal{M} \} +1}} + \sum\limits_{k = \mathrm{max \{ \mathcal{M} \} + 1 }}^{K} \frac{F_k}{f_{\mathrm{max}}} \nonumber \\
      &= C_{\mathrm{max \{ \mathcal{M}_2 \} }} + \sum\limits_{k = \mathrm{max \{ \mathcal{M}_2 \}  }}^{K} \frac{F_k}{f_{\mathrm{max}}} \nonumber \\
      &= T_K^{(M,\mathcal{M}_2)} = T_K^{(M+1,\mathcal{M}_3)},
\end{align}
where the right hand-side of the inequality in  \eqref{eq:infInequalityM} holds because for the  
$(M,\mathcal{M})$-broken queue $I_{\mathrm{max \{ \mathcal{M} \} } } > C_{\mathrm{max \{ \mathcal{M} \} +1}}$ is satisfied and the last equality is true by the definition of the $(M,\mathcal{M}_2)$-broken queue and the fact that it has the same last break as the $(M+1,\mathcal{M}_3)$-broken queue.
Note that if a queue consists only of $M=2$ sub-queues the same technique cannot be used because there is only $1$ break meaning that the defined sets cannot be defined and the bandwidth reallocation cannot be utilized.

\bibliographystyle{IEEEtran}
\bibliography{bib.bib}

\end{document}